\documentclass[letterpaper, 10 pt, journal, twoside]{ieeetran} 
\IEEEoverridecommandlockouts

\usepackage{ifthen}
\newboolean{mark_changes}
\setboolean{mark_changes}{false} % true to show changes. false to only show the new version

\usepackage[absolute,overlay]{textpos}
\usepackage{graphics} % for pdf, bitmapped graphics files
\usepackage{epsfig} % for postscript graphics files
\usepackage{cite} %To compress multiple cites
\usepackage{amsmath}
\usepackage{amssymb} % assumes amsmath package installed
\usepackage{mathtools}
\usepackage{esvect}
\usepackage{graphicx}
\usepackage{makecell}
\usepackage[pdfa]{hyperref}
\usepackage{comment}
\usepackage{multirow}
\usepackage{ctable}
\usepackage[normalem]{ulem}
\usepackage[bottom]{footmisc} %force footnote to be at the bottom

\graphicspath{ {./img/} }

\DeclareMathOperator{\sigm}{sigm}
\DeclareMathOperator{\diag}{diag}

\newcommand{\norm}[1]{\left\lVert#1\right\rVert}
\usepackage{color}

\ifthenelse{\boolean{mark_changes}}
{
    \newcommand{\revision}[1]{\textcolor{blue}{#1}} % highlight changes
}
{
    \newcommand{\revision}[1]{#1} % don't highlighting changes
}

\title{Perception-Aware Perching on Powerlines with Multirotors}

\author{J. L. Paneque,$^{1}$ J.R. Mart\'{\i}nez-de Dios,$^{1}$ A. Ollero,$^{1}$ D. Hanover,$^{2}$ S. Sun,$^{2}$ A. Romero,$^{2}$ and D. Scaramuzza$^{2}$
\thanks{Manuscript received: September, 9, 2021; Revised November, 10, 2021; Accepted January, 4, 2022.}%Use only for final RAL version
\thanks{This paper was recommended for publication by Editor P. Ponds upon evaluation of the Associate Editor and Reviewers' comments.
This work was supported by the European Union’s Horizon 2020 Research and Innovation Programme under grant agreement No. 871479 (AERIAL-CORE), by the European Research Council (ERC) under grant agreements No. 788247 (GRIFFIN) and No. 864042 (AGILEFLIGHT), by the National Centre of Competence in Research (NCCR) Robotics through the Swiss National Science Foundation (SNSF), and by the Spanish Ministerio de Universidades FPU Program.
}%Use only for final RAL version
\thanks{$^{1}$J. L. Paneque, J.R. Mart\'{\i}nez-de Dios and A. Ollero are with the GRVC Robotics Lab Sevilla, Universidad de Sevilla, Seville, Spain
        {\tt\footnotesize (email: jlpaneque@us.es, jdedios@us.es, aollero@us.es)}}%
\thanks{$^{2} $D. Hanover, S. Sun, A. Romero and D. Scaramuzza are with the Robotics and Perception Group, Department of Informatics, University of Zurich, and Department of Neuroinformatics, University of Zurich and ETH Zurich, Zürich, Switzerland
        {\tt\footnotesize (email: djhanove@umich.edu; sun@ifi.uzh.ch; roagui@ifi.uzh.ch; davide.scaramuzza@ieee.org)}}%
\thanks{Digital Object Identifier (DOI): doi.org/10.1109/LRA.2022.3145514}

}

\begin{document}
\maketitle
%\thispagestyle{empty}
%\pagestyle{empty}
% Comment or remove these lines for final RAL version.

\begin{textblock*}{20cm}(0.5cm,27cm) % {block width} (coords) 
  \scriptsize © 2022 IEEE.  Personal use of this material is permitted.  Permission from IEEE must be obtained for all other uses, in any current or future media, including reprinting/republishing this material for advertising or promotional purposes, creating new collective works, for resale or redistribution to servers or lists, or reuse of any copyrighted component of this work in other works.
\end{textblock*}

\begin{abstract}
Multirotor aerial robots are becoming widely used for the inspection of powerlines. To enable continuous, robust inspection without human intervention, the robots must be able to perch on the powerlines to recharge their batteries. Highly versatile perching capabilities are necessary to adapt to the variety of configurations and constraints that are present in real powerline systems. This paper presents a novel perching trajectory generation framework that computes perception-aware, collision-free, and dynamically-feasible maneuvers to guide the robot to the desired final state. Trajectory generation is achieved via solving a Nonlinear Programming problem using the Primal-Dual Interior Point method. The problem considers the full dynamic model of the robot down to its single rotor thrusts and minimizes the final pose and velocity errors while avoiding collisions and maximizing the visibility of the powerline during the maneuver. The generated maneuvers consider both the perching and the posterior recovery trajectories. The framework adopts costs and constraints defined by efficient mathematical representations of powerlines, enabling online onboard execution in resource-constrained hardware. The method is validated on-board an agile quadrotor conducting powerline inspection and various perching maneuvers with final pitch values of up to 180\textdegree. 
%The developed code is available online at: \url{https://github.com/grvcPerception/pa_powerline_perching}
% We release our code fully open-sourced. % Remove this for space, since the link is already highlighted below as an URL
\end{abstract}
% Keywords appear just beneath the abstract. Use only for final RAL version. 
%\begin{IEEEkeywords}
%Aerial Systems: Applications; Aerial Systems: Perception and Autonomy; Constrained Motion Planning; Vision-Based Navigation; Collision Avoidance
%\end{IEEEkeywords}

\section*{Supplementary Material}

Video: \url{https://youtu.be/JsPavnsfpbk}

Code: \href{https://github.com/grvcPerception/pa_powerline_perching}{github.com/grvcPerception/pa\_powerline\_perching}

\section{Introduction}
\label{sec:intro}

% \subsection{Motivation} % Remove this for space (still have the contribution later)
% What is the problem? Why is it important?
\IEEEPARstart{T}{he} use of multicopters in the inspection of hazardous industrial environments (e.g., nuclear plants, steel mills, or power lines) provides significant opportunity to drastically reduce the risk of human injury on the job-site.
%In the US, over 1,200 bridges need to be inspected per day to ensure that the structural integrity of the bridge is maintained over time~\cite{cherni}.
The US power system consists of nearly 160,000 miles of high-voltage power lines, and millions of low-voltage power lines and distribution transformers, which connect 145 million customers~\cite{hoff_2016}.
Inspections of this magnitude require extensive manpower and work hours in highly dangerous environments. 
Leveraging robust autonomous robots for inspection of infrastructures could improve throughput of these inspections, thus reducing the possibility of failure. %which could lead to the loss of human life. 
%According to recent estimates, the overall drone inspection & monitoring market is projected to grow from USD 9.1 billion in 2021 to USD 33.6 Billion by 2030~\cite{globe_newswire_2021}. %\angel{This is very nice intro! one thing though is that AERIALCORE money comes from Europe, would it be maybe beneficial for us to put some European numbers as well?}

\begin{figure}
    \centering
    \includegraphics[width=\linewidth]{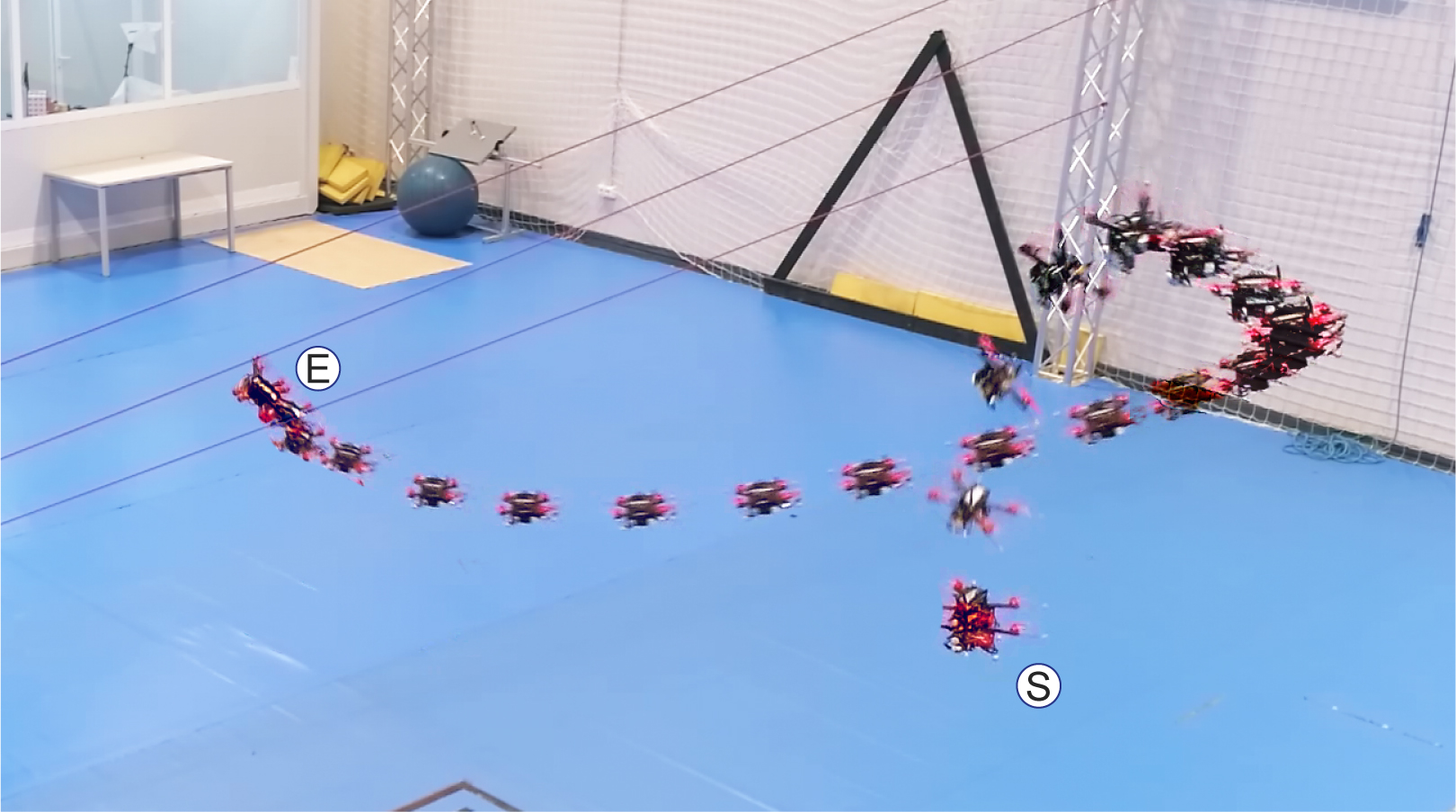}
    \caption{A quadrotor performing a perception-aware perching maneuver, maximizing the visibility of the line during the whole trajectory. \revision{The maneuver starts at the bottom-right corner of the image (S), and progresses first upwards to adjust its orientation, and then towards the objective line (E).}
    }
    \label{fig:introduction}
    %\vspace{-5mm}   
\end{figure}

% Why is the problem hard? What makes it challenging?
In order to enable these opportunities, multicopters must be able to operate in uncertain highly-cluttered environments, varying environmental conditions, and with limited onboard energy. 
The multicopter is then responsible for estimating its state via onboard sensors, calculating agile trajectories which maximize inspection coverage, and executing dynamic maneuvers near crowded, safety critical infrastructures. 
The powerline inspection task offers the possibility of perching directly on the powerlines to recharge onboard battery systems via wireless charging \cite{kitchen2020batcharge}.
This has the potential to improve efficiency of powerline inspection drones over manned inspection, but requires the unmanned system to perch on the line. The challenge is then to design algorithms which can identify powerlines and relevant obstacles in the observable space, plan a trajectory that satisfies the dynamic constraints of the multirotor and avoids obstacles while keeping the goal point in view (see Fig. \ref{fig:introduction}), and execute the trajectory in a potentially windy or rainy environment. 

% How far has existing work come? Why hasn't the problem been solved? What is the stumbling block? What is the frontier?
Most systems perch on vertical walls by directing the robot towards them and adjusting their angle during the maneuver \cite{mellinger2012trajectory,thomas2016aggressive}. Existing systems for perching on cables rely on approaching while hovering the objective line either from the top \cite{hang2019b} or the bottom \cite{thomas2015visual} assuming there is enough space to do so.
%Perching on actual powerlines is complex due to obstacle density and  geometric irregularities. 
Actual powerlines can come in many different configurations \revision{where this may not be possible,} necessitating that any planning algorithm must be able to account for these configurations directly and plan accordingly. 
Simply installing multiple perching devices will not suffice because it limits the weight and efficiency of the platform. 

% What does our paper contribute? What is the key idea? What is the magic trick? What is the new insight or technique that enables us to advance the frontier?
\subsection{Contribution}
This paper presents a novel perching trajectory generation framework for powerlines, which produces highly versatile, agile, collision-free, and dynamically-feasible maneuvers to guide the robot to a desired pose at zero velocity while maximizing the visibility of the objective powerline. It is based on a Nonlinear Programming (NLP) optimization problem that uses a nonlinear quadrotor model formulated down to the rotor thrust level. The NLP uses a simplified mathematical representation that efficiently represents the powerlines (with negligible errors) as concatenations of segments and the robot as an ellipsoid with three different radii.

The paper has four main contributions: 1) a highly-versatile perception-aware agile perching trajectory generation method based on NLP; 2) a general mathematical modeling for collision avoidance and perception awareness near powerlines; 3) experimental validation in different agile  maneuvers including extreme 180\textdegree {} perching; and 4) the developed code of the trajectory generator, which is released to the robotics community. We demonstrate the efficacy of the proposed method onboard a quadrotor, first using the mathematical modeling inside an onboard Nonlinear Model Predictive Controller (NMPC) to perform inspection flights, and then using the developed framework to generate perching maneuvers in three different lines, which were followed with final pitch values of 20\textdegree, 80\textdegree, and 180\textdegree. 

The paper is organized as follows: Section \ref{sec:rw} summarizes the main works in the topics addressed in the paper. Section \ref{sec:problem} presents the problem formulation for powerline perching. Section \ref{sec:modeling} describes the proposed mathematical modeling. Section \ref{sec:traj-generation} presents the method for perching trajectory generation. Section \ref{sec:experimental} provides an experimental validation of the developed work. Finally, Section  \ref{sec:conclusion} concludes the paper and highlights the main future research steps.

\section{Related Work}
\label{sec:rw}

Prior works on multirotor perching have usually focused on the problem of agile perching on walls. First, authors in \cite{mellinger2012trajectory} performed perching trajectories by compounding multiple linear control modes that did not guarantee the feasibility of the maneuver. Later, the work in \cite{thomas2016aggressive} addressed this problem by planning for dynamically feasible maneuvers before their execution, which is also the case in the proposed method. Other works on perching on walls have usually focused on the design of the perching mechanism \cite{daler2013fibre,tsu2015door}. Recently, a visual perching for walls was presented in \cite{mao2021perching}, were the authors use a combination of Apriltags and Visual Inertial Odometry to perch on walls without a Motion Capture System. All these systems usually attach on walls by colliding with them with some final velocity, while the proposed one reaches the perching state at zero (or desired) velocity. On the other hand, works for multirotor perching on cylinders (which can include powerlines) are usually limited to reaching the desired spot at a hover state and attaching to the cylinder using a gripper \cite{thomas2015visual}, \cite{popek2018agrasp}. A heterogeneous perching platform is proposed on \cite{hang2019b}, which can rest or perch in a variety of different situations, provided they are reached from above. For powerlines this is not always the case, since they appear in  many different configurations where hanging from the upmost line can lead to touching the others. Recently, \cite{Yu2020perching} presented a quadrotor with upside-down perching capabilities by using bi-directional thrusts. In contrast, the proposed system is able to generate perching trajectories that take multirotors to any (feasible) desired perching state, including upside-down, without the use of bi-directional thrusts.

During perching maneuvers, it is important to keep visibility of the final objective, either if it is a landing area, a cylinder, or a powerline. Traditional methods such as Image-Based Visual Servoing \cite{thomas2015visual} enforce this naturally by formulating the control law in the image space coordinates. However, aggressive perching maneuvers may not always have the final spot inside its Field of View (FOV), especially if the robot's camera is not located at the perching mechanism. Authors in \cite{falanga2018PAMPC} presented a perception-aware NMPC for multirotors which uses additional error terms in its cost function to keep visibility of a desired object while tracking a trajectory. The work in \cite{jacquet2020PCMPC} then proposed to keep different targets inside the multirotor's FOV by formulating their visibility as constraints inside an NMPC controller. Later, authors in \cite{greeff2020PaFlatMPC} developed a probabilistic constraint to keep the number of successfully matched visual landmarks over a minimum threshold during a flight, including the multirotor's gimbal in the modeled dynamics of their NMPC. We take inspiration in these works and include perception awareness in the generated perching trajectories by formulating novel costs and constraints designed for lines and segments instead of point landmarks, so their perception can be considered in the computation of the perching maneuver.

\section{Problem Formulation}
\label{sec:problem}

The objective of powerline perching is to guide an aerial robot to a desired final pose with zero velocity, where it can grip to a powerline. This has to be done while avoiding  collisions and maximizing the visibility of the powerline during the trajectory. The basic scenario is composed of a set of powerlines, not necessarily parallel, with several tens of meters of length and at a certain height. 

Powerlines follow catenary equations, whose use for costs and constraints formulation in NLP systems would result in very inefficient implementations. In our approach, to overcome this issue, we adopt a mathematical model that approximates catenary shapes as concatenations of segments. Segments can be very efficiently integrated in NLPs both for measuring robot-powerline distances (to ensure collision-free maneuvers) and also for estimating the powerline visibility from the robot camera. Powerlines can be represented by several segments to provide an accurate representation, and there are already algorithms to perform the approximation  \cite{phillips1968algorithms}. For instance, we measured that a real powerline of 185m could be modeled by 15 segments with a mean length of 12m and a mean error of 1'5 cm with respect to it. Moreover, an average perching maneuver will only involve 1 or 2 of these segments per line.

While perching maneuvers usually end when the robot reaches the objective pose, it may happen that the perching device fails to attach to the powerline. In that case, the system must be able to recover to a safe state, while still avoiding the powerlines. Our NLP framework is used to also compute this recovery trajectory, appended to the perching maneuver.

\section{Mathematical modeling}
\label{sec:modeling}

\subsection{Nomenclature}

In this work, we follow standard conventions and denote scalar magnitudes with lowercase $s$, vectors with bold lowercase $\mathbf{v}$, and matrices with bold uppercase $\mathbf{M}$. We also make use of different reference frames, all defined with uppercase $F$ and with an orthonormal basis $\{\mathbf{x}_F,\mathbf{y}_F,\mathbf{z}_F\}$.

We represent translations between two coordinate frames as vectors $\mathbf{p}_{F_1 F_2} \in \mathbb{R}^3$, such that a vector $\mathbf{v}_{F_2} \in \mathbb{R}^3$ is expressed in $F_1$ as: $\mathbf{v}_{F_1} = \mathbf{p}_{F_1 F_2} + \mathbf{v}_{F_2}$. For rotations, we use unit quaternions $\mathbf{q} \in \, ${SO}(3), which can be expressed in different frames as $\mathbf{q}_{F_1} =  \mathbf{q}_{F_1 F_2} \odot \mathbf{q}_{F_2}$, where $\odot$ denotes the Hamilton product between two quaternions. Finally, we define the rotation of a vector $\mathbf{v} \in \mathbb{R}^3$ by a quaternion with the following abuse of notation: $\mathbf{q} \odot \mathbf{v} \coloneqq \mathbf{q} \odot \left[0, \mathbf{v}\right]^T$.

\subsection{Multirotor Dynamics Model}

Similarly to \cite{sun2021comparative,jacquet2020PCMPC}, we model our multirotor robot as a rigid body of mass $m$ and diagonal moment of inertia matrix $\mathbf{J}$, with nominal dynamics $\dot{\mathbf{x}}$ down to their second order derivatives. The robot is actuated by the thrusts $\boldsymbol{\gamma} \in \mathbb{R}^4$ of four individually-controllable rotors, i.e., $\boldsymbol{\gamma} = \left[\gamma_1,\gamma_2,\gamma_3,\gamma_4\right]^T $. Typically, the individual rotor thrusts $\gamma_i$ are used as the control inputs of the dynamic system and then are translated into desired rotational speeds for the motors using a simple algebraic relation \cite{Faessler17ral}. However, the rotors actually behave as a first-order system with a time constant of several ms, which means they cannot change their thrust instantaneously as demanded by the controller. This effect is of high importance when generating perching trajectories, which demand fast deceleration and rotation before the end of the trajectory. Assuming instantaneous thrust dynamics potentially leads to  generating dynamically unfeasible maneuvers that cannot be followed by the multirotor. To solve this, we model the inputs of the system  as the desired constant thrust derivatives  $\boldsymbol{u} \in \mathbb{R}^4$ and include the thrusts $\boldsymbol{\gamma}$ as part of the state of the system, similarly to \cite{jacquet2020PCMPC}. This ensures continuity in the required actuations and allows to include the physical limits of the rotor angular accelerations and decelerations in the NLP framework.

The 17-dimensional robot state space is then defined as:
\begin{equation}
    \label{eq:state-space}
    \dot{\mathbf{x}} 
    = 
    \begin{bmatrix} 
        \dot{\mathbf{p}}_{W B} \\ 
        \dot{\mathbf{q}}_{W B} \\  
        \dot{\mathbf{v}}_W \\ 
        \dot{\boldsymbol{\omega}}_B \\ 
        \dot{\boldsymbol{\gamma}} 
    \end{bmatrix} 
    = 
    \begin{bmatrix} 
        \mathbf{v}_W \\ 
        \mathbf{q}_{W B} \odot \left[0, {\boldsymbol{\omega}_B}^T/2 \right]^T \\  
        \frac{1}{m} \mathbf{q}_{W B} \odot \mathbf{\Gamma}_B + \mathbf{g}_W \\ 
        \mathbf{J}^{-1}\left( \mathbf{M} \boldsymbol{\gamma} - \boldsymbol{\omega}_B \times \mathbf{J}\boldsymbol{\omega}_B \right) \\ 
        \boldsymbol{u} 
    \end{bmatrix}
    \in \mathbb{R}^{17},
\end{equation}
where $\textbf{p}_{W B}$ and $\textbf{q}_{W B}$ are the position and orientation of the robot's body frame $B$ w.r.t. the world frame $W$, and ${\mathbf{v}}_W$ and ${\boldsymbol{\omega}}_B$ $\in \mathbb{R}^3$ are the linear and angular velocities of the multirotor robot, measured in global and body axes of the robot respectively. Vector $\mathbf{g}_W \in \mathbb{R}^3$ denotes the acceleration due to gravity in global axes. The vector $\mathbf{\Gamma}_B \in \mathbb{R}^3$  encodes the collective thrust of the motors in the body axes, where in our case all 4 motors are directed to $B_z$. Finally, $\mathbf{M}\in \mathbb{R}^{3 \times 4}$ is the thrust allocation matrix that converts the current rotor thrusts into body torques in $B$:
\begin{equation}
    \label{eq:thrust-alloc-mat}
    \mathbf{\Gamma}_B = 
    \begin{bmatrix}
    0 \\
    0 \\
    \sum\boldsymbol{\gamma}
    \end{bmatrix}
    \quad
    \mathbf{M} = 
    \begin{bmatrix}
    \mathbf{r_y}^T \\
    \mathbf{-r_x}^T \\
    \kappa \mathbf{r_d}^T
    \end{bmatrix},
\end{equation}
where $\mathbf{r_x}$ and $\mathbf{r_y} \in \mathbb{R}^4$ are the rotor displacements in $B_x$ and $B_y$, $\kappa$ is the rotor drag torque constant, and $\mathbf{r_d} \in {\{-1,1\}}^4$ are the individual rotor spin directions, where ${r_d}_i = -1$ for counter-clockwise direction and ${r_d}_i = 1$, otherwise.

\subsection{Segment collision avoidance}

First, the robot-powerline collision is modeled assuming there is only one straight powerline. Then, the model is extended to powerlines composed of several segments.

Let the robot's body be represented as an ellipsoid with principal axes $\{B_x,B_y,B_z\}$ and principal radii $\boldsymbol{\delta}$. Assume there is only one straight powerline whose radius is summed in $\boldsymbol{\delta}$. The parametric equation of the line is given by $\mathbf{o}_W + \tau \mathbf{l}_W$,
%\noindent 
where $\textbf{o}_W$ and $\textbf{l}_W \in \mathbb{R}^3$ are the origin and direction vectors of the line, and $\tau$ is a parameter. We can transform the line to the body frame $B$ and scale it with $\boldsymbol{\delta}_B$:
\begin{equation} \label{eq:rescale}
    \mathbf{o}_{\Breve{B}} = \mathbf{\Delta}_B \mathbf{q}_{B W} \odot \left(\mathbf{o}_W - \mathbf{p}_{W B}\right), \, \mathbf{l}_{\Breve{B}} = \mathbf{\Delta}_B \mathbf{q}_{B W} \odot \mathbf{l}_W,
\end{equation}

\noindent where $\boldsymbol{\Delta}_B$ = $\diag\left(1/\{\delta_x, \delta_y, \delta_z\}\right)$. Note that if $\boldsymbol{\delta}$ is defined in a different frame than $B$, \eqref{eq:rescale} can be rearranged using further transformations until $\textbf{o}_W, \textbf{l}_W$ are in its same frame.

Now that the line lies in the scaled reference frame of the ellipsoid, ensuring that there is no intersection between them is equal to proving that the distance from the line to the origin of that reference frame is higher than 1. The squared point-line distance formula from the origin of $\Breve{B}$ gives:
\begin{equation} \label{eq:pldistance}
    \frac{\norm{\mathbf{o}_{\Breve{B}} \times \mathbf{l}_{\Breve{B}}}^{2}}{\norm{\mathbf{l}_{\Breve{B}}}^{2}} > 1,
\end{equation}
\noindent which can finally be simplified by using $\norm{\mathbf{a} \times \mathbf{b}}^{2} = \norm{\mathbf{a}}^{2}\norm{\mathbf{b}}^{2} - ( \mathbf{a} \cdot \mathbf{b} )^2$:
\begin{equation} 
    \label{eq:constraint}
    \left(\norm{\mathbf{o}_{\Breve{B}}}^{2} -1\right)\norm{\mathbf{l}_{\Breve{B}}}^{2} - (\mathbf{o}_{\Breve{B}} \cdot \mathbf{l}_{\Breve{B}} )^2 > 0
\end{equation}

%  The satisfaction of \eqref{eq:constraint} determines if there are any or no intersections with the line. % Removed for redundancy

When working with real powerlines, we need to use more than one straight segment to approximate the curved shape of the powerline in the maneuver's surroundings. In that case, the collision constraint \eqref{eq:constraint} could be activated outside of its corresponding segment and interfere with the perching maneuver. Thus, we need to extend this constraint such that it is not activated outside of its segment's area of effect.

The minimum value of \eqref{eq:constraint} is reached whenever $\mathbf{o}_{\Breve{B}}=\mathbf{0}$ or $\mathbf{o}_{\Breve{B}}\parallel \mathbf{l}_{\Breve{B}}$.
%, due to the properties of the dot product. 
In these cases, the value will be $-\norm{\mathbf{l}_{\Breve{B}}}^2$. In \eqref{eq:rescale}, $\mathbf{l}_{\Breve{B}}$ is a matrix multiplied by a rotated unit direction vector. The maximum possible value of $\norm{\mathbf{l}_{\Breve{B}}}$ appears when $\mathbf{q}_{B W} \odot \mathbf{l}_W$ is aligned with the principal eigenvector of $\boldsymbol{\Delta}_B$, and corresponds to $\lambda_{1}(\boldsymbol{\Delta}_B)$, which is the maximum eigenvalue of $\boldsymbol{\Delta}_{{B}}$. 
\revision{Thus, we can force the collision avoidance function \eqref{eq:constraint} to be always positive by summing $\lambda^2_{1}(\boldsymbol{\Delta}_B)$ to it. We define the following function to add that value whenever the robot is outside of the segment's surroundings:}
%We can use this to define a function $\kca(\mathbf{x})$ which is summed to \eqref{eq:constraint} and which forces that the collision avoidance constraint is always satisfied when the distance to the center of the segment is higher than half its length: 
\begin{equation} \label{eq:augmentation}
    {k(\mathbf{x})} = \lambda^2_{1}(\boldsymbol{\Delta}_B)\sigm( ( \norm{\mathbf{p}_{W B} - \mathbf{o}_{W}} ) )),
\end{equation}
\noindent where $\norm{\mathbf{p}_{W B} - \mathbf{o}_{W}}$ is the distance from the robot to the segment's center, and $\sigm \left( \cdot\right)$ can be any sigmoid function that is scaled and translated such that $k$ is 0 when this distance is \revision{lower} than half the segment's length \revision{plus the highest radii in $\mathbf{\delta}_B$}, and is $\lambda^2_{1}(\boldsymbol{\Delta}_B)$ 
%when the robot is completely outside of it
\revision{otherwise, since then the robot will never collide with the segment, even when intersecting with its corresponding line}. We chose the $\arctan$ function since it is available in two of the main NLP code-generation frameworks \cite{Houska2011acado,andersson2019casadi}, while others such as $\tanh$ are only available in \cite{andersson2019casadi}. 
Finally, summing \eqref{eq:augmentation} into \eqref{eq:constraint} ensures that the resulting collision avoidance constraint is only activated in the surroundings of %its corresponding
\revision{the} segment:
\begin{equation}
     h_{ca}\left(\textbf{x}\right) \coloneqq \left(\norm{\mathbf{o}_{\Breve{B}}}^{2} -1\right)\norm{\mathbf{l}_{\Breve{B}}}^{2} - (\mathbf{o}_{\Breve{B}} \cdot \mathbf{l}_{\Breve{B}} )^2 + k(\mathbf{x}) > 0
\end{equation}
\subsection{Segment perception awareness}

Following the previous approach, we first assume there is only one straight powerline in the scenario, then extend to the segment-based case. Let $C$ be the reference frame of a camera mounted on the robot. The position and orientation of $C$ are given by $\textbf{p}_{W C}$ and $\textbf{q}_{W C}$, which are computed from the robot's current body pose and a fixed transformation $T_{BC}=\{\textbf{p}_{B C},\textbf{q}_{B C}\}$. A line is expressed in frame $C$ as:
\begin{equation}
    \mathbf{o}_{C} = \mathbf{q}_{C W} \odot \left( \mathbf{o}_W - \mathbf{p}_{W C}\right), \quad  \mathbf{l}_{C} = \mathbf{q}_{C W} \odot \mathbf{l}_W
\end{equation}

%We then redefine the line by its correspondent Pl\"ucker coordinates, i.e. the normal of the plane that intersects with such line and the origin, and the direction vector of the line: $\{\mathbf{n}_C, \mathbf{l}_C \}$, with $\mathbf{n}_C = \mathbf{o}_{C} \times \mathbf{l}_{C}$.
We then redefine the line by its Pl\"ucker coordinates, i.e. the normal of the plane that intersects with it and the origin, and its direction vector: $\{\mathbf{n}_C, \mathbf{l}_C \}$, with $\mathbf{n}_C = \mathbf{o}_{C} \times \mathbf{l}_{C}$.

Assume a classical pinhole camera model with parameters $\{f_x, f_y, c_x, c_y \}$. For brevity, assume the pixel coordinates are centered at the optical axis (i.e., $c_x=c_y=0$). The transformation of the direction vector onto the 3D image frame $I$ is given by $\mathbf{l}_I = \mathbf{K}_P \mathbf{l}_C$, with $\mathbf{K}_P=\diag\left(f_x,f_y,1\right)  \in \mathbb{R}^{3\times 3}$ being the intrinsic camera matrix. Similarly, the vector $\mathbf{n}_C$ is transformed onto the image coordinates as $\mathbf{K}_L \mathbf{n}_C$, with $\mathbf{K}_L = \diag\left(f_y,f_x,f_x f_y\right)  \in \mathbb{R}^{3\times 3}$.
The point-line reprojection error for a given 2D image point $\mathbf{m}$ is  \cite{fu2020plvins}:
\begin{equation} \label{eq:perception-cost}
    \widetilde{r}\left(\textbf{x}\right) = \frac{\underline{\mathbf{m}}^T \mathbf{n}_I }{\sqrt{ n_{I,x}^2 + n_{I,y}^2 }},
\end{equation}
\noindent where $\underline{\mathbf{m}} \in \mathbb{R}^3$ is the 2D point in homogeneous coordinates.

As stated in Section \ref{sec:rw}, it is convenient to keep the tracked objects (either points, lines, or other shapes) as close as possible to the center of the image. This allows the robot to focus on such objects and avoid losing track of them, potentially improving the accuracy of the object's localization overtime (which is especially important for perching maneuvers and inspection tasks). We can achieve this by choosing $\underline{\mathbf{m}}^T = \mathbf{e}_z = \left[0,0,1\right]^T$ and minimizing \eqref{eq:perception-cost} for it.

However, there are two ways in which this function can be driven to zero: by having the line centered in front of the camera, and by doing so behind the camera. The second case is undesirable, since for a single pinhole camera this means the system may not see the line. We need an additional constraint to ensure the line is centered in front of the camera.
To define this constraint, we first obtain two new vectors:
\begin{equation} \label{eq:newvecs}
\mathbf{p}^{2D}_I = \mathbf{n}_I \times \left(\mathbf{e}_z \times \mathbf{n}_I\right), \quad \mathbf{d}^{3D}_I = \mathbf{l}_I \times \mathbf{n}_I
\end{equation}

\begin{figure}
% \vspace{1mm}
    \centering
    \includegraphics[width=\linewidth, height=3.5cm]{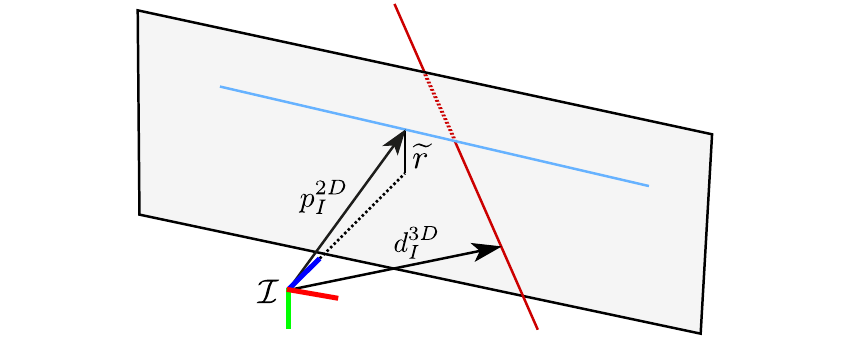}
    \caption{Visualization of proposed mathematical modeling for \eqref{eq:perception-cost} and \eqref{eq:newvecs}.}
    \label{fig:perception-constraints}
    % \vspace{-3mm}
\end{figure}

The vector $\mathbf{p}^{2D}_I  \in \mathbb{R}^3$ is directed to the closest point of the line from the center of the image when the line is in 2D normalized image coordinates (Fig. \ref{fig:perception-constraints}). Conversely, the vector $\mathbf{d}^{3D}_I \in \mathbb{R}^3$ is the closest point from the line to the origin of $I$ when the line is in 3D unnormalized image coordinates.

As stated before, we are interested in keeping the 2D line as close as possible to the center of the image, which is the same as keeping $\mathbf{p}^{2D}_I$ as parallel as possible to $\mathbf{e}_z$. Since the vector $\mathbf{p}^{2D}_I$ marks where is the nearest point of the line from the center of the image, if we obtain its intersection with the 3D line we can recover its sign and force that is positive, thus having the line in front of the camera when minimizing \eqref{eq:perception-cost}. The result of solving such intersection is:
\begin{equation}
    \mathbf{p}^{3D}_I = \frac{\mathbf{p}^{2D}_I }{d^{3D}_{I,z}}, \quad {p}^{3D}_{I,z} = \frac{ n_{I,x}^2 + n_{I,y}^2 }{d^{3D}_{I,z}}
\end{equation}

Note that the sign of ${p}^{3D}_{I,z}$ is determined by the sign of $d^{3D}_{I,z}$ as its numerator will always be $\geq0$. Thus, forcing the line to be centered in front of the camera is equivalent to forcing the following line cheirality (i.e., side) constraint: 
\begin{equation} \label{eq:line-cheirality-constraint}
    h_{lc}\left(\textbf{x}\right) \coloneqq {d}^{3D}_{I,z} > 0
\end{equation}

We now extend the given formulation to work with segments, by defining a third constraint that is complementary to  \eqref{eq:line-cheirality-constraint}.
Let $\mathbf{e1}_I$ and $\mathbf{e2}_I  \in \mathbb{R}^3$ be the two endpoints of the inspected segment in the 3D image coordinates. These points lie in the same line as $\mathbf{p}^{3D}_I$, so the dot product between $\left(\mathbf{p}^{3D}_I - \mathbf{e1}_I\right)$ and $\left(\mathbf{p}^{3D}_I - \mathbf{e2}_I\right)$ is negative whenever $\mathbf{p}^{3D}_I$ is located between $\mathbf{e1}_I$ and $\mathbf{e2}_I$. This serves to create a constraint to keep $\mathbf{p}^{3D}_I$ between the endpoints of the segment. If the line is centered, this means at least half of the image will contain the segment. The proposed segment visibility constraint is thus formulated as:
\begin{equation}
    \label{eq:segmment-visivility-constraint}
    h_{sv}\left(\textbf{x}\right) \coloneqq (-1)\left(\mathbf{p}^{3D}_I - \mathbf{e1}_I\right) \cdot \left(\mathbf{p}^{3D}_I - \mathbf{e2}_I\right) > 0
\end{equation}

\section{Perching Trajectory Generation}
\label{sec:traj-generation}

\subsection{Optimization Problem Formulation}

We model the perching maneuver generation as a discrete-time multiple-shooting NLP problem sampled in $N$ shooting points over a non-fixed time horizon $T$, which is also an optimization variable of the problem:
\begin{subequations} \label{eq:op}
\begin{flalign}
    \label{eq:op-cost} \min_{ \begin{matrix} \boldsymbol{u}_0...\boldsymbol{u}_{N-1} \\T \end{matrix}} \quad & \sum_{k=0}^{N} \norm{\mathbf{\widetilde{y}}_k}^2_{\mathbf{Q}_k} & \\
    \label{eq:op-xinit} \textrm{s.t.} \quad \quad \; & \mathbf{x}_0 = \mathbf{x}_{init} & \\ 
    \label{eq:op-time} & T_{min} \leq T \leq T_{max} \\
    \label{eq:op-dyn} & \mathbf{x}_{k+1} = \textbf{f}\left(\mathbf{x}_k,\boldsymbol{u}_k\right) & \forall k \in \left[0,N-1\right] \\
    \label{eq:op-z} & z_{min} \leq {p}_{W B,z} & \forall k \in \left[0,N\right] \\
    \label{eq:op-motors} & 0 \leq \boldsymbol{\gamma} \leq \gamma_{max} & \forall k \in \left[0,N\right] \\
    \label{eq:op-actuations} & u_{min} \leq \boldsymbol{u} \leq u_{max} & \forall k \in \left[0,N-1\right] \\
    \label{eq:op-cheirality} & 0 < h_{lc}\left(\textbf{x}_{k}\right)  & \forall k \in \left[0,N\right] \\
    \label{eq:op-visibility} & 0 < h_{sv}\left(\textbf{x}_{k}\right) & \forall k \in \left[0,N\right] \\
    \label{eq:op-avoidance} & 0 \leq h_{ca,i}\left(\textbf{x}_{k}\right) & 
        \begin{matrix}
        \forall k \in \left[0,N\right] \\
        \forall i \in \left[0,N_L-1\right]
        \end{matrix}
\end{flalign}
\end{subequations}

%\angel{You really need to make clear in this optimization problem formulation that you are using it to plan/generate the perching trajectory. These formulations are usually quickly identified with control methods, that's why I think it's very important that you introduce what you use it for in a more clear way.}\julio{I have made it more clear now}

The problem \eqref{eq:op} is built as follows: \eqref{eq:op-cost} is the cost function to minimize, including final and running terms; \eqref{eq:op-time} are the limits of the total maneuver time $T$; \eqref{eq:op-dyn} are the dynamics of the system (see next paragraph); \eqref{eq:op-z} is the allowed minimum height; \eqref{eq:op-motors},\eqref{eq:op-actuations} are the constraints for the motor thrusts and their derivatives; \eqref{eq:op-cheirality},\eqref{eq:op-visibility} are the line cheirality \eqref{eq:line-cheirality-constraint} and segment visibility \eqref{eq:segmment-visivility-constraint} constraints for the objective line; and \eqref{eq:op-avoidance} are the line avoidance constraints defined by \eqref{eq:constraint} and \eqref{eq:augmentation} for all the present segments.

We implement the variable time horizon by modeling the system dynamics \eqref{eq:op-dyn} using a Runge-Kutta4 integration of the state space, scaling its derivative \eqref{eq:state-space} by the total time $T$ and using an integration step of $1 / N$ seconds. Since problems where the total maneuver time is an optimization variable suffer from bad linearization characteristics, we chose the ForcesPRO framework \cite{FORCESPro} with \cite{FORCESNLP} as the NLP solver, which embedded a linear system solver with high numerical stability. Convergence was typically achieved between 100 and 1000 iterations of its Nonlinear Primal-Dual Interior-Point method, depending on the complexity of the required maneuver. This was a feasible requirement since each maneuver is only computed once before execution. A further analysis of this is provided in Section \ref{sec:experimental}.

The cost function \eqref{eq:op-cost} consists of a set of errors $\mathbf{\widetilde{y}}_k$ dependent on the states and the inputs of the system, and weighted by a diagonal matrix $\mathbf{Q}_k$ for every shooting node. Different values of $\mathbf{\widetilde{y}}_k$ are used to model the terminal and running costs (reference frames omitted for brevity):
\begin{equation}
    \mathbf{\widetilde{y}}_k = \begin{cases}
        \left[ {\left[\boldsymbol{\gamma}_{k}\frac{T}{N} + \mathbf{u}_k\frac{T^2}{2N^2}\right]}^T \, \mathbf{{w}}^T_k \, {\widetilde{r}\left(\textbf{x}_{k}\right)} \right]^T &  k \in \left[0,N-1\right] \\
        \left[ \mathbf{\widetilde{p}}^T_{k} \, \mathbf{\widetilde{q}}^T_k \, \mathbf{\widetilde{v}}^T_k  \, \mathbf{\widetilde{w}}^T_k \right]^T & k = N 
    \end{cases}
\end{equation}

It can be seen that $\mathbf{\widetilde{y}}_k \in \mathbb{R}^8$ for the running cost and $\mathbf{\widetilde{y}}_k \in \mathbb{R}^{12}$ for the terminal cost. The running cost minimizes the integral of the motor thrusts (which is $\int_0^{\frac{T}{N}} \left( \boldsymbol{\gamma} + \boldsymbol{u}t \right)d_t = \boldsymbol{\gamma}\frac{T}{N} + \mathbf{u}\frac{T^2}{2N^2}$ ) , as well as the angular velocities $\mathbf{{w}}_k$ of the robot and the reprojection error $\widetilde{r}_k$ of the objective line. The terminal cost minimizes the position and orientation error $\mathbf{\widetilde{p}}_{k} , \mathbf{\widetilde{q}}_k$ as well as the final linear and angular velocity errors $\mathbf{\widetilde{v}}_k , \mathbf{\widetilde{w}}_k$ at the desired perching state $\mathbf{x}_{perch}$.
The constraints \eqref{eq:op-cheirality}-\eqref{eq:op-avoidance} are always present during the whole horizon prediction. However, while satisfying constraint \eqref{eq:op-avoidance} is critical to avoid collisions with powerlines, doing so for the perception constraints is not practical, since the camera may be mounted at a different place from the perching mechanism, and thus will not see the powerline at the final part of the maneuver. To solve this, we model the constraints \eqref{eq:op-cheirality},\eqref{eq:op-visibility} as soft constraints with exponentially decaying costs, such that they are negligible at the end of the trajectory. We do the same for the cost of $\widetilde{r}_k$. Finally, notice that since the lines are in global axes (i.e., mapped with any state estimator that tracks their position in $W$), the NLP does not need them to be inside the camera's FOV in order to work.

\begin{figure}
\vspace{1mm}
    \centering
    \includegraphics[]{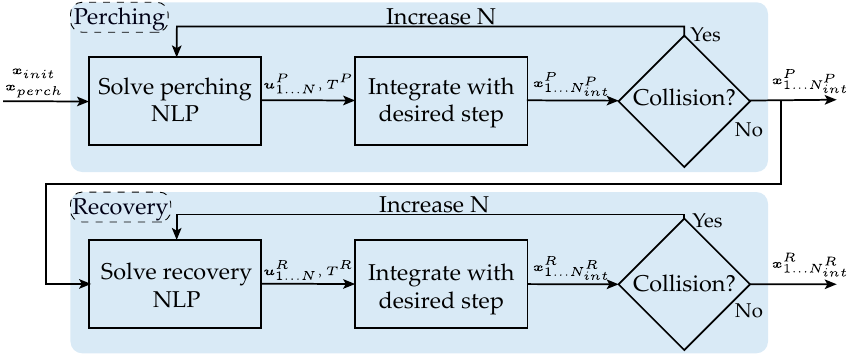}
    \caption{Developed procedure for perching+recovery trajectory integration.}
    \label{fig:perching-recovery-integration}
    % \vspace{-1mm}
\end{figure}

% Conversely, if the camera was mounted at the perching mechanism, the costs would be exponentially increasing to only enforce perception at the end of the maneuver. 

\subsection{Trajectory and Recovery Integration}

We are interested in perching trajectories that can recover the robot to a safe position without any collisions even if the perching mechanism fails. To do so, we use the same optimization problem from \eqref{eq:op} with different cost values (and without perception costs and constraints) to generate a recovery trajectory that starts right after the perching trajectory finishes (see Fig. \ref{fig:perching-recovery-integration}): First, we solve \eqref{eq:op} to compute the perching trajectory. Second, we integrate its result with a finer resolution (we used 1ms for the proposed experiments) using a Runge-Kutta4 integration scheme. Third, we check the integrated trajectory for any possible collisions between shooting nodes that could have not been detected when solving the NLP. If necessary, we can solve \eqref{eq:op} again with a higher $N$ using the current solution as a warm start (we found $N=30$ is usually good enough for trajectories of several meters and $T\leq5$ s). Finally, we  use the end of the perching trajectory as the beginning of the recovery and solve \eqref{eq:op} for it, also integrating its result afterwards and adding it after the integrated perching maneuver.

The resulting trajectory is continuous for the whole maneuver, intrinsically leads to a safe recovery if the perching is not completed, and is dynamically feasible.

\begin{figure}[t]
% \vspace{3mm}
    \centering
    \includegraphics[width=\linewidth, height=4.9cm]{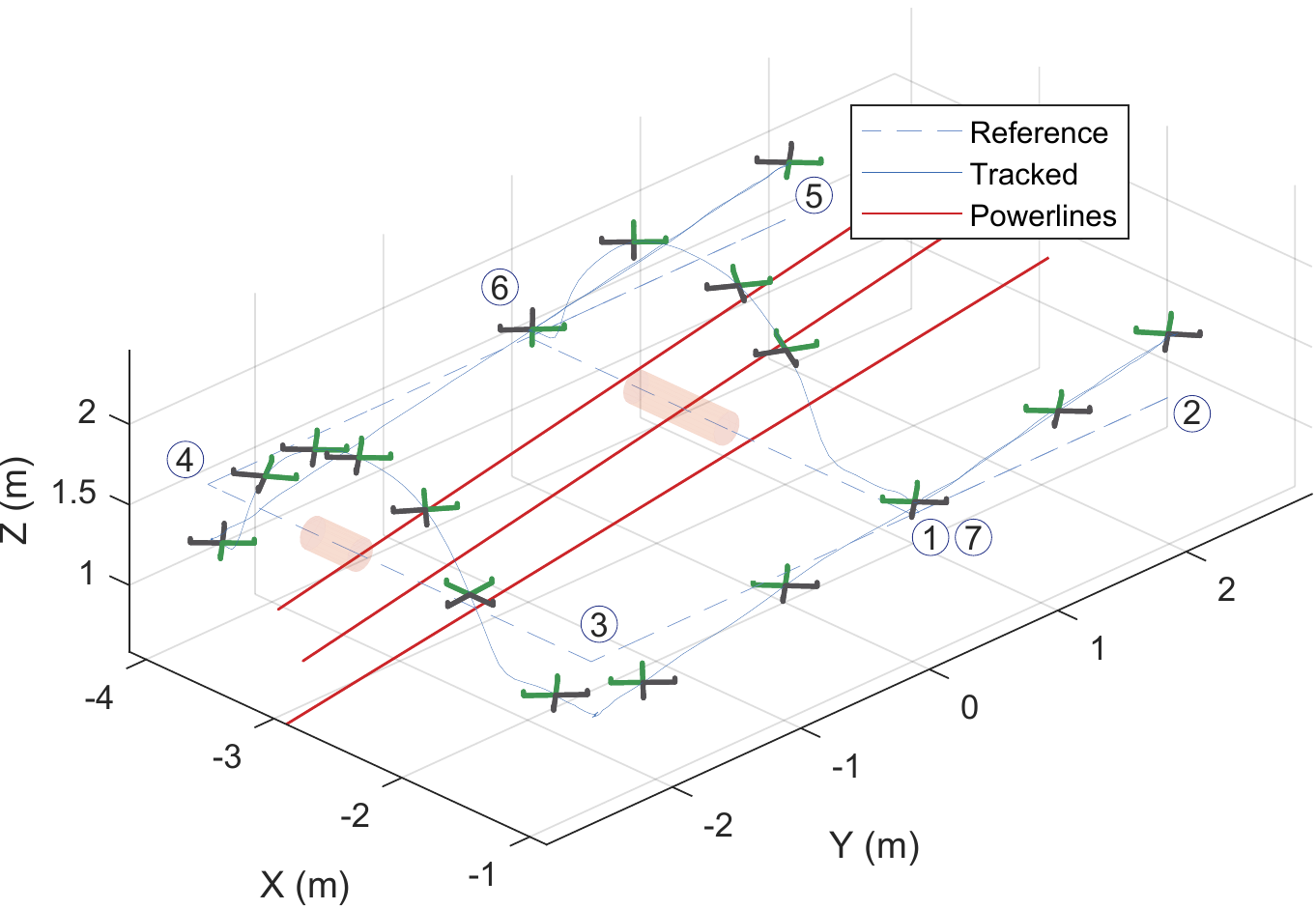}
    \caption{Actual robot trajectory performed by the resulting NMPC in an experiment where the provided reference had two collisions (red zones). The front of the robot is colored in green. The powerlines appear in red. The numbers denote the commanded waypoints.}
    \label{fig:inspection1}
\end{figure}

%\angel{To this point I would have expected more details about the algorithm itself, what is the prediction sample time (time between samples), if you solve your optimization problem once per trajectory or multiple times, etc} \julio{I added them now.}

\section{Experimental Validation}
\label{sec:experimental}

The proposed method was evaluated onboard a custom quadrotor platform developed at the Robotics and Perception Group (RPG) of the University of Zurich, with a weight of 0.8 kg and a thrust-to-weight ratio of 4:1. A Radix FC board was used as the low level flight controller, and a NVIDIA  Jetson TX2 as the main onboard computer. The robot was equipped with a Realsense D435i camera in its front face.
%and 9 cm below its reference frame $B$. 
The state estimates of the quadrotor are given by an optical tracking system running at 120Hz, while the positions of the mockup powerlines are obtained 
%\revision{at 10Hz} 
by \cite{fu2020plvins}, a visual-inertial state estimator which maps point and line features \revision{with 10Hz visual feedback (enough for mapping static lines), running in the CPU of the system}. 
It was adapted to map red lines by only using the images' red channel in its line search module. 
%\revision{No GPU adaptation of the code was performed.}

The developed method was implemented in C++ following the scheme in Fig. \ref{fig:perching-recovery-integration}, and then interfaced as a ROS node. Different experiments were performed with the developed NLP with and without perception awareness. To remove the perception awareness, it is only necessary to disable constraints \eqref{eq:op-cheirality} and \eqref{eq:op-visibility}, and remove \eqref{eq:perception-cost} from the
NLP, which is done by setting their correspondent weights to zero. The control actuations of the quadrotor during flight are computed by a NMPC controller from RPG described at \cite{sun2021comparative}. The controller runs at 100Hz using a Real-Time Iteration (RTI) scheme \cite{diehl2006fast} and is then cascaded with a high-frequency L1 adaptive controller \cite{drew2021adaptiveNMPC} that corrects for disturbances such as aerodynamic drag or model inaccuracies.

\begin{figure}[t]
%  \vspace{3mm}
    \centering
    \includegraphics[width=\linewidth, height=4.9cm]{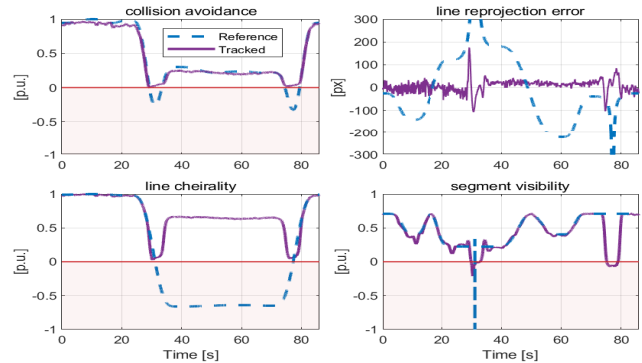}
    \caption{Collision avoidance constraint \eqref{eq:op-avoidance}, line reprojection error \eqref{eq:perception-cost}, and line cheirality \eqref{eq:op-cheirality} and segment visibility \eqref{eq:op-visibility} constraints during the experiment in Fig. \ref{fig:inspection1}. The constraint violation zone is marked in red.}
    \label{fig:inspection2}
\end{figure}

Finally, in this work we are not interested in the development of a specific perching mechanism, but rather in providing the optimal positioning that such mechanisms would require for the perching to happen. Thus, we focus the experiments on the maneuvering itself and always recover the quadrotor to a safe position after reaching the perching state.

\subsection{Inspection experiments}
\label{sec:ex-inspection}

First, we validate the mathematical modeling from Section \ref{sec:modeling} decoupled from the perching trajectory generation system. To do so, we incorporate constraints \eqref{eq:op-cheirality}-\eqref{eq:op-avoidance} and cost \eqref{eq:perception-cost} into the onboard NMPC controller described in \cite{sun2021comparative}, without making  use yet of the proposed NLP trajectory generation system. The resulting controller is validated through missions where the robot performs powerline inspection (see Fig. \ref{fig:introduction} for a visual clue of the line setup). The robot is commanded to follow straight lines  between a set of waypoints which is intentionally thought to lead the robot to collide with the powerlines and to maintain them far from the center of the camera. The resulting NMPC controller follows the given trajectories adapting its yaw and height to avoid collisions while maximizing the visibility of the required powerline. Fig. \ref{fig:inspection1} shows that the trajectory actually performed by the robot successfully avoided the two potential collisions in contrast to the waypoint trajectory. Fig. \ref{fig:inspection2} compares the reprojection error between the commanded trajectory and the one actually followed by the NMPC controller, showing a mean improvement of 500\%, which involves 100 pixels. The perception and collision avoidance constraints are satisfied during the whole flight, except for the segment visibility constraint during brief instants. This exhibits the advantages of including the proposed modeling in the NMPC controller.
Another advantage is that the line positions can be updated online since the NMPC executes at real time.
However, notice that the controller still requires a feasible trajectory or reference to follow. For agile perching, this can not be simply the desired pose or a hover-to-hover minimum snap trajectory. The proposed NLP solves the trajectory generation and is validated in Section \ref{sec:ex-perching}. Moreover, since the NLP already accounts for collision avoidance and perception awareness, the controller will not need to include these, so other trajectory-tracking controllers with lower computational demand could be used \cite{sun2021comparative}.

More than 20 experiments were performed providing the robot with various trajectories and powerline configurations, and the resulting NMPC always achieved similar results. All these experiments were performed with up to three segments and with a 100 Hz RTI control rate with the only requirement of formulating all the included constraints as soft ones, 
whose costs are started at zero and are slowly increased at the beginning of the flights.
If more than one line should be inspected at the same time, one could append more costs and constraints for each line (increasing the computational cost), or track the centroid of the lines as an intermediate solution.

\subsection{Perching experiments}
\label{sec:ex-perching}

\begin{figure}[t]
%  \vspace{3mm}
    \centering
    \includegraphics[width=\linewidth, height=70mm]{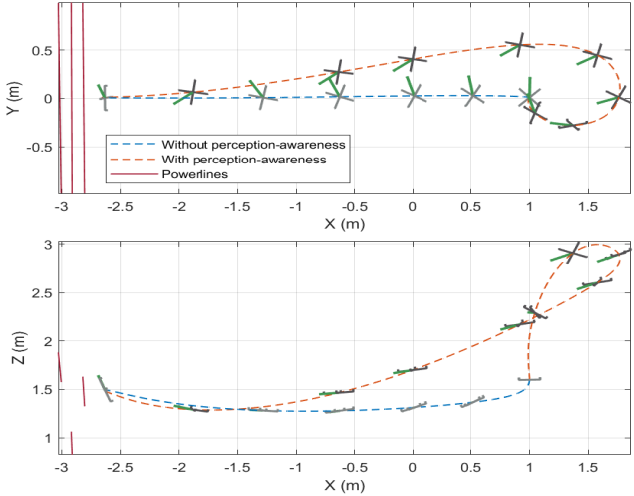}
    \caption{Comparison of two perching maneuvers with and without perception awareness. The perception-aware maneuver  (orange)  is followed by the black quadrotor, while the other (blue) is followed by the grey quadrotor. The green lines show the orientation of the robot's camera at each instant.}
    \label{fig:perching1}
\end{figure}

We now validate the whole proposed perching trajectory generation system in different maneuvers that are computed and executed onboard the robot. We assume the perching end-effector is installed at the bottom of the quadrotor as in \cite{Yu2020perching}. We add an additional degree of freedom to the end-effector, and assume its yaw orientation can be controlled, so we can better illustrate the effect of perception awareness.
%To better illustrate the effect of perception awareness, we add an additional degree of freedom to the end-effector, and assume its yaw orientation can be controlled. 

\begin{figure}[t]
%  \vspace{3mm}
    \centering
    \includegraphics[width=\linewidth,height=5.2cm]{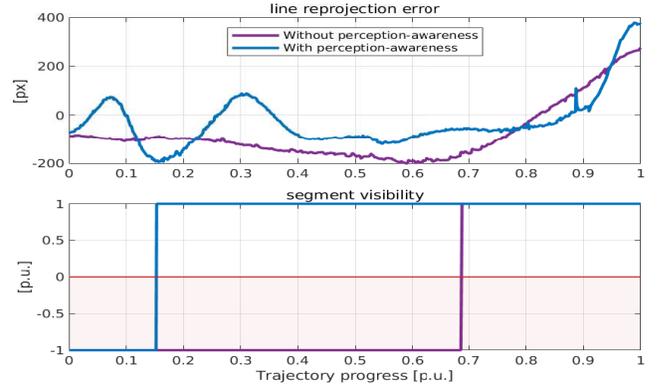}
    \caption{ Line reprojection error \eqref{eq:perception-cost} and segment visibility constraint \eqref{eq:op-visibility} during both maneuvers in Fig. \ref{fig:perching1}. The obstacle avoidance \eqref{eq:op-avoidance} and line cheirality \eqref{eq:op-cheirality} constraints are always satisfied for both maneuvers and thus are not shown. The constraint violation zone is marked in red.}
    \label{fig:perching2}
\end{figure}

Three powerlines were set up with different inclinations (see Fig. \ref{fig:introduction}). The robot was set to hover in front of them with its camera parallel to the lines. The robot first performs a perching maneuver to reach the closest line at 80\textdegree {} without including perception awareness. The robot is able to follow the trajectory, reach the perching pose with zero velocity, and recover to a safe position afterwards. Then, the maneuver is computed and executed for the same end pose, but including perception awareness. Fig. \ref{fig:perching1} shows both trajectories from the same initial point. By performing an initial correction, the quadrotor is able to reach the same end pose while keeping the powerline centered in its camera during most of the trajectory. Fig. \ref{fig:perching2} shows the evolution of the reprojection error and the segment visibility constraint during the maneuver, which was more favorable in the perception-aware case. A comparison of final position and orientation errors within multiple flights is given later on Table \ref{tab:exp-results}.

\begin{figure}
\vspace{0.5mm}
    \centering
    \includegraphics[width=\linewidth, height=6.5cm]{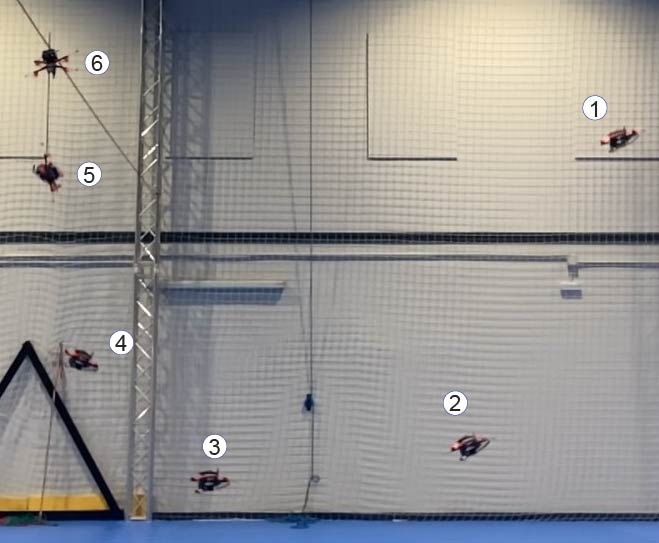}
    \caption{Tracked perching maneuver where the robot finishes upside-down. The numbers denote the evolution of the trajectory.}
    \label{fig:perching3}
    % \vspace{-4mm}
\end{figure}

After showing the capabilities for perception-awareness, we test its functioning in extreme perching maneuvers, where the drone stops completely upside down at the perching pose. We note that this has currently only been done with quadrotors with bi-directional thrust capabilities \cite{Yu2020perching}. We set up the bottom powerline at a height of 3'7 m, leaving roughly 2'5 m of operation  for the quadrotor in the Z axis (its minimum allowed height is of 0.8 m). The robot is set to hover in front of the line and then computes and executes the required perching \revision{maneuver}. %Fig. \ref{fig:perching3} 
%shows the trajectory followed by the quadrotor.
Its tracking can be seen in Fig. \ref{fig:perching3}.
%The resulting generated \revision{trajectory} can be followed by the robot even in extreme cases such as perching upside down. 
Moreover, the posterior recovery of the system is also accomplished without reaching the demanded minimum height. 
%\revision{In this case the computed trajectory with perception-awareness was identical to the one without it, since the already restrictive demanded maneuver leaves no margin to reorientate the robot's front during it.}
\revision{In this case the effect of the perception awareness becomes negligible unless its corresponding costs are significantly increased, resulting in a divided maneuver that first moves towards the line while keeping it in view and then only performs perching when the perception costs decay. This is an expected outcome since the already restrictive demanded maneuver leaves no margin to reorientate the robot's front during it. Thus, this result is not further analyzed.}
%In this case the effect of perception-awareness is negligible due to the restrictive demanded trajectory, and thus it is not shown.

\begin{table}[]
% \vspace{4mm}
    \centering
    \begin{tabular}{c c c c c}
    \specialrule{.1em}{.05em}{.05em} 
    Maneuver & $\mathbf{{p}}_{RMSE}$ (cm) & $\mathbf{{q}}_{RMSE}$ (\textdegree) & $t_{p}$ (ms) & $t_{r}$ (ms) \\ \hline
    80 \textdegree      & 3.1 & 2.0 & 432 & 221 \\
    80 \textdegree + PA & 3.4 & 2.6 & 1526 & 584  \\
    20 \textdegree      & 2.7 & 2.1 & 457 & 351 \\ 
    20 \textdegree + PA & 4.8 & 7.2 & 1893 & 461 \\
    180 \textdegree     & 6.9 & 3.0 & 532 & 1214  \\
    \specialrule{.1em}{.05em}{.05em} 
    \end{tabular}
    \caption{Position and pitch-roll rmse of different tracked perching trajectories, and computational time to produce them. }
    \label{tab:exp-results}
    % \vspace{-4mm}
\end{table}

Finally, we analyze the average performance of the whole system in a set of different perching experiments. We repeat the experiments presented in Figs. \ref{fig:perching1} and \ref{fig:perching3}, and add an additional experiment where the robot performs a perching maneuver to the top line in Fig. \ref{fig:perching1}, at a pitch angle of 20\textdegree, with and without perception awareness. For 5 experiments of each maneuver, we compute the mean RMSE position and orientation errors at the perching point, as well as the time required for computing the perching and recovery maneuvers. Table \ref{tab:exp-results} summarizes these results. We observed that in general the trajectories could be tracked with fairly low position error with the exception of the upside-down perching. This could potentially be improved with a finer tuning of the onboard controller, though it can be compensated by having a certain degree of tolerance in the perching mechanism. On the other hand, we found that the performance of perception-aware maneuvers was highly dependent on the starting position with respect to the objective line (all performed maneuvers started at the same relative position to the whole setup). For example, the 20\textdegree PA maneuver was harder to compute and follow since it was started from a lower altitude than the objective line, while this was not the case in the one with 80\textdegree . This opens future research on how to compute the optimal starting point for a perching maneuver.

% \begin{figure}
% \vspace{2.5mm}
%     \centering
%     % \includegraphics[width=\linewidth, height=5.5cm]{img/ex5.jpeg}
%     \includegraphics[trim=20 10 10 25, clip, width=\linewidth, height=5.0cm]{img/ex5_new.jpg}
%     \caption{Tracked perching maneuver where the robot finishes upside-down. The numbers denote the evolution of the trajectory.}
%     \label{fig:perching3}
%     % \vspace{-4mm}
% \end{figure}

\section{Conclusions and Future Work}
\label{sec:conclusion}

In this work we presented a novel perching trajectory generation framework which generates highly versatile perching trajectories that satisfy collision avoidance with powerlines and maximize their visibility during flight. 
The efficacy of our method was demonstrated on a set of real world experiments onboard a computationally limited quadrotor. 
We show that the quadrotor is capable of executing the perching trajectory with minimal tracking error and complete obstacle avoidance, even at very high angles of attack. 
If the perching mechanism were to fail, our algorithm provides a fail safe trajectory such that the drone automatically recovers and maintains flight.
Additionally, we show that our formulation is capable of running inside an onboard controller in real time, providing it with capabilities for inspection of powerlines, avoiding collisions with them and ensuring that the inspected line is kept in view at all times. 
In the future, we want to explore how the starting point of a perching trajectory impacts its performance, and how to utilize this information to increase the likelihood of a successful perching. We also hope to explore how multiple cameras or sensors can be taken into account simultaneously into the same perching maneuver, combining their potential at the parts of the maneuver where they best suit for.

\section*{Acknowledgments}
The authors would like to thank Thomas Laengle for his help in the setup of the quadrotor used for the experiments, Philipp Foehn and Elia Kaufmann for their helpful insights on the first version of the manuscript, and  Leonard Bauersfeld and V\'{\i}ctor Valseca for their help in the preparation of the media material.

\bibliographystyle{IEEEtran}
\bibliography{powerlinePerchingRAL}

\end{document}